\documentclass[letterpaper, 10 pt, conference]{ieeeconf}  

\usepackage[colorlinks,urlcolor=blue,linkcolor=blue,citecolor=blue]{hyperref}
\usepackage[most]{tcolorbox}
\usepackage{graphicx} 
\usepackage{cite} 
\usepackage{amsmath}
\IEEEoverridecommandlockouts                              
\title{\LARGE \bf{Neurosymbolic AI approach to Attribution in Large Language Models}
}

\begin{document}

\author{
    Deepa Tilwani \\
    \textit{AI Institute} \\
    University of South Carolina \\
    Columbia, SC, USA \\
    \and
    Revathy Venkataramanan \\
    \textit{AI Institute} \\
    University of South Carolina \\
    Columbia, SC, USA \\
    \and
    Amit P. Sheth \\
    \textit{AI Institute} \\
    University of South Carolina \\
    Columbia, SC, USA
}

\maketitle
\thispagestyle{empty}
\pagestyle{empty}

\begin{abstract}

Attribution in large language models (LLMs) remains a significant challenge, particularly in ensuring the factual accuracy and reliability of the generated outputs. Current methods for citation or attribution, such as those employed by tools like Perplexity.ai and Bing Search-integrated LLMs, attempt to ground responses by providing real-time search results and citations. However, so far, these approaches suffer from issues such as hallucinations, biases, surface-level relevance matching, and the complexity of managing vast, unfiltered knowledge sources. While tools like Perplexity.ai dynamically integrate web-based information and citations, they often rely on inconsistent sources such as blog posts or unreliable sources, which limits their overall reliability. We present that these challenges can be mitigated by integrating Neurosymbolic AI (NesyAI), which combines the strengths of neural networks with structured symbolic reasoning. NesyAI offers transparent, interpretable, and dynamic reasoning processes, addressing the limitations of current attribution methods by incorporating structured symbolic knowledge with flexible, neural-based learning. This paper explores how NesyAI frameworks can enhance existing attribution models, offering more reliable, interpretable, and adaptable systems for LLMs.
\end{abstract}

\maketitle
Large language models (LLMs) have become essential tools in applications ranging from conversational AI to knowledge-intensive tasks. However, they face a significant challenge in attribution—citing sources for generated content—which is crucial for ensuring factual accuracy and building user trust, especially in high-stakes domains like healthcare, law, and education. LLMs often ``hallucinate" \cite{Xu2024HallucinationII}, generating fabricated information that cannot be traced back to credible sources \cite{peskoff-stewart-2023-credible}, complicating the attribution process \cite{Li2023ASO}. This becomes especially apparent in scenarios involving information-seeking and knowledge-based question-answering, where users depend on LLMs for expert insights \cite{Malaviya2023ExpertQAEQ}. For instance, consider a student using an AI tool to write a research paper. The student may unknowingly commit plagiarism if the AI provides a verbatim paragraph from a copyrighted article without proper citation. This risks academic penalties and could lead to legal issues for the institution. In another case, an AI-generated report might include citations after nearly every sentence, some pointing to unreliable blog posts. Such over-citation with poor-quality sources clutters the report and diminishes its credibility.

\begin{figure}[h!]
    \centering
    \includegraphics[width=\linewidth]{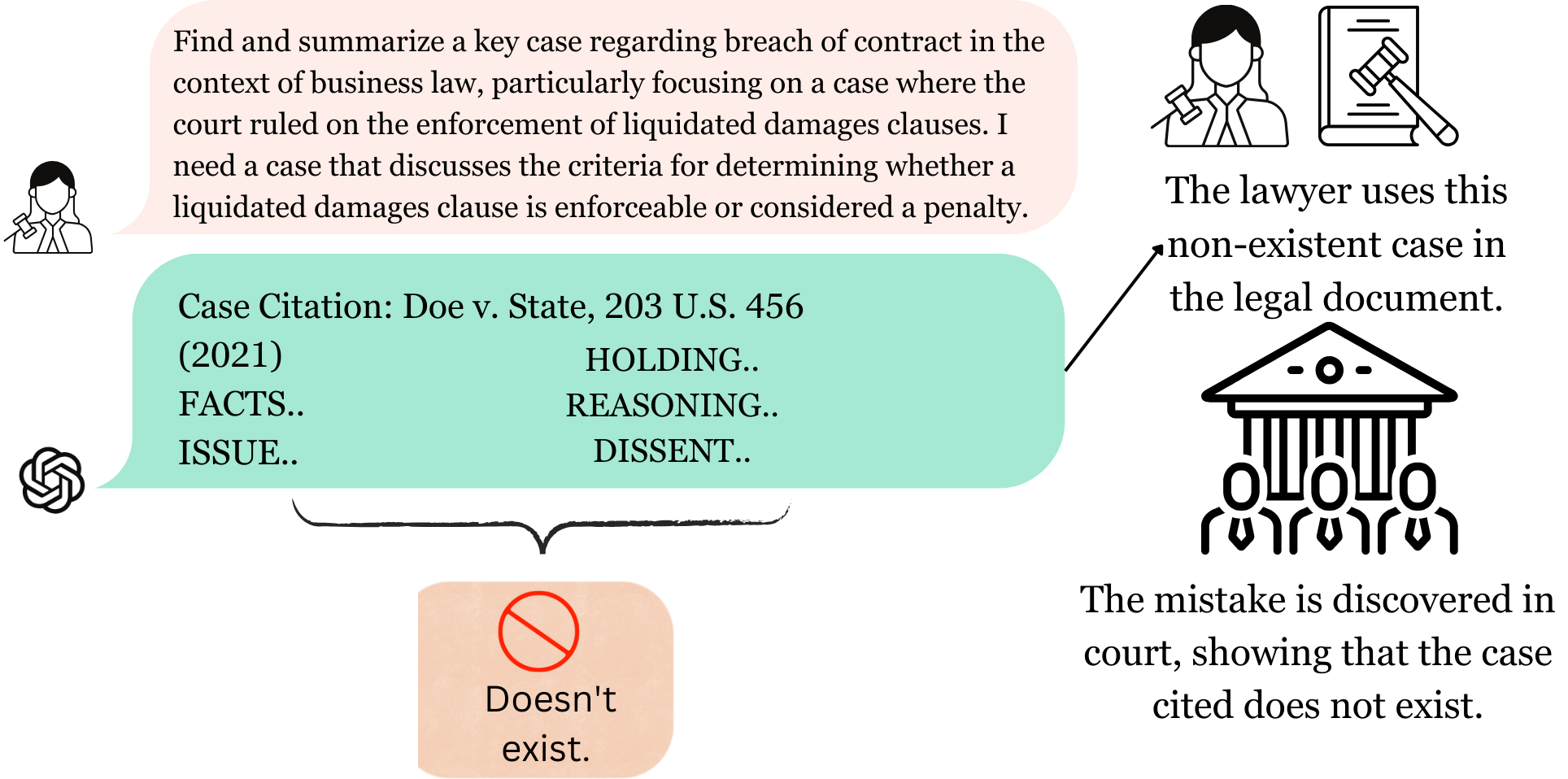}
    \caption{A lawyer drafting a legal brief with the help of AI is misled into citing a fabricated case, ``Doe v. State (2021)." The error is later discovered, damaging the lawyer's credibility and jeopardizing the case (for visualization purposes, not a real case)}
    \label{fig:lawyer_ai_error}
\end{figure}

In critical domains such as health and law, the consequences of incorrect attribution can be very severe\cite{jayakumar-etal-2023-large}. The reliance on LLM-generated content by legal professionals, highlighted by The New York Times, illuminates the pitfalls when these models produce content that lacks proper verification. In a notable incident, a lawyer used an AI assistant to draft a legal brief and was misled into citing a fabricated case in an airline lawsuit\footnote{https://tinyurl.com/45cvrv6x} \footnote{https://tinyurl.com/2s3ewbtk}. This oversight required the attorney to apologize and accept sanctions, damaging their credibility and jeopardizing the case. Another example, as shown in \autoref{fig:lawyer_ai_error}, is an illustrative case not drawn from actual events. It highlights the potential for errors when utilizing AI in high-stake legal settings. In the healthcare domain, producing unreliable outputs could lead to adverse outcomes. It is mandatory for these models to provide explanations for their generated content and support those explanations with sources to which the explanations can be traced back. Such accountability could prevent troublesome outcomes.

The vast and unfiltered datasets intensify these challenges used to train LLMs, making it difficult to trace specific information back to reliable sources accurately. Current attribution methods often lack precision and thoroughness, resulting in overcitation, inadequate sourcing, and inherent data biases \cite{Liu2023EvaluatingVI}. Legal issues have emerged, with lawsuits against companies like OpenAI, Microsoft, and Stability AI, alleging the use of copyrighted materials without proper attribution. These claims highlight potential copyright infringements and contribute to a growing erosion of user trust\footnote{https://tinyurl.com/4u6nwvh6},\footnote{ https://tinyurl.com/56rswja8}.

Methods like retrieval-augmented generation (RAG) for citations \cite{Tilwani2024REASONSAB, li2023towards} and post-generation citation \cite{bohnet2022attributed} have been developed to improve accuracy, they still fail to provide fully reliable, traceable, and understandable sources. Retrieval-based models, which aim to fetch references, suffer from limitations such as outdated information and lack of real-time updates, reducing their reliability \cite{Asai2024ReliableAA}. Without a structured approach to ensure factual accuracy, LLMs deliver misleading information and fail to adapt to evolving knowledge. This raises a critical question:

\begin{tcolorbox}[colback=gray!10, colframe=black!60, boxrule=0.5pt, arc=4pt, left=6pt, right=6pt, top=6pt, bottom=6pt, box align=center, width=\columnwidth]
    \centering
    \emph{“How can we ensure that LLM-generated content is reliably and transparently attributed to verifiable sources in real-time, particularly for critical applications in domains such as health and law?”}
\end{tcolorbox}

\begin{figure*}[ht!] 
    \centering
    \includegraphics[width=\textwidth]{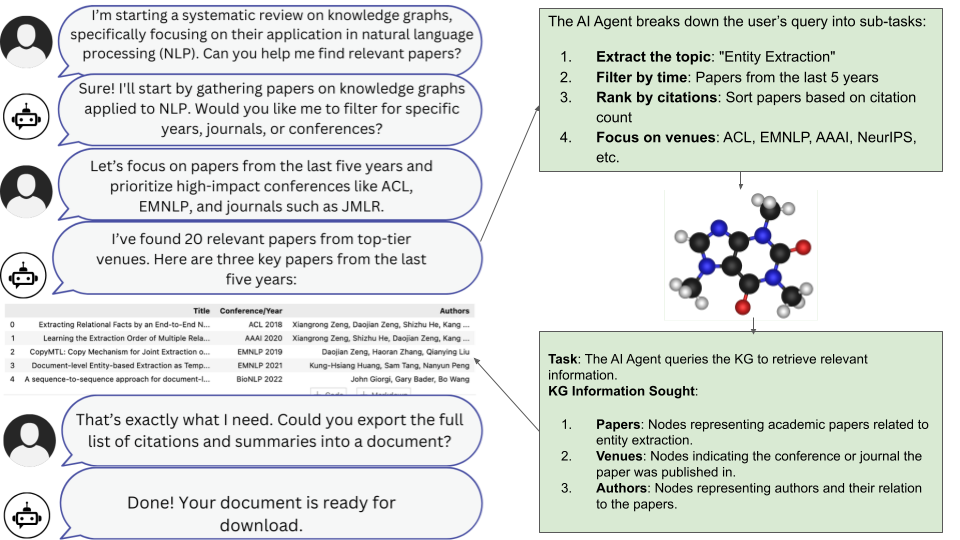} 
    \caption{State-of-the-art LLMs could not handle attribution reliably and accurately. This figure demonstrates how an AI agent, integrated with a knowledge graph, systematically retrieves relevant academic papers on a given topic, filters by time and venue, and ranks by citation count, ensuring the accuracy and reliability of the sources provided to the user.}
    \label{fig:intro2}
\end{figure*}

Attribution requires precision—the ability to back every claim with a reliable source—and comprehensiveness—ensuring no generated content is left uncited or poorly sourced. The unfiltered nature of LLM training data amplifies these challenges, leading to ambiguous and incorrect attributions.

Existing solutions, such as Perplexity.ai and BingChat, rely on LLMs to provide contextual responses by integrating real-time web search results, but they face significant limitations \cite{Liu2023EvaluatingVI}. These systems often pull information from sources of varying quality, including blog posts, and struggle with relevance matching and biases in source selection. Specifically, they tend to favor more popular sources over reliable but less frequently referenced ones, leading to potential misinformation. Our approach directly addresses this bias in source selection by prioritizing verified, high-quality databases through symbolic reasoning, thus improving the reliability and interpretability of attributions in LLMs.

Using the Neurosymbolic AI (NesyAI) framework, which combines symbolic reasoning with the statistical capabilities of neural networks, aims to create more robust, reliable, and interpretable attribution systems by integrating symbolic rules with neural adaptability to validate and generate accurate citations dynamically. The \autoref{fig:intro2} illustrates how a NesyAI agent can systematically query a knowledge graph (KG) to retrieve papers, filter by venue, time, and citation count, and provide users with reliable, well-organized references. This approach could significantly improve attribution mechanisms in LLMs by ensuring that generated outputs are grounded in verifiable data from trusted sources.

\section{Neurosymbolic AI as a Framework for Attribution}

NesyAI combines neural network-based perception with symbolic reasoning to achieve both large-scale data processing and structured reasoning \cite{Sheth2023NeurosymbolicA}. This hybrid approach mirrors how humans process information—integrating raw sensory input with abstract knowledge. NesyAI enables explicit tracking of reasoning steps through symbolic representations, such as KGs \cite{Gaur2023BuildingTN}. In doing so, NesyAI offers a natural framework to address the key challenges of attribution by integrating flexible learning with grounded \cite{goonmeetGrounding}, verifiable knowledge. As highlighted in this framework, KGs offer a clear pathway to attribution\cite{Adam2024TraceableLV}. KGs can maintain a dynamic yet traceable repository of factual information, ensuring that every claim generated by an LLM is grounded in a reliable source. These graphs can be updated in real-time, allowing the system to remain relevant even as knowledge evolves.

\section{How Neurosymbolic AI Solves the Attribution Problem?}

\subsection{\textbf{Structured Knowledge Representation}}
The use of structured symbolic knowledge, such as KGs and ontologies, can dramatically improve how attribution is handled. KG represent unambiguous, dynamic, and rationalizable knowledge—the facets needed to anchor claims and enhance the precision of attributions. In NesyAI, KGs provide explicit, interpretable structures that ensure every piece of generated information has a verifiable source. By integrating these graphs into the attributed response generation process, LLMs could align their outputs with verifiable, pre-approved sources, reducing the possibility of hallucination. For example, by associating claims generated by the neural components of an LLM with pre-existing entries in a KG, we can create a link between dynamic content generation and static, verifiable data sources. This approach helps ensure that every statement made by an LLM can be traced to a specific, reliable piece of knowledge, eliminating the ambiguity in model-generated content.

\subsection{\textbf{Memory Structures and Temporal Reasoning}}
NesyAI frameworks fundamentally address the limitations of fixed memory systems by incorporating dynamic memory structures capable of tracking and integrating real-time updates. The dynamic memory layer in NesyAI would interact with external data sources that receive continuous updates (e.g., real-time databases, APIs, or sensor data) to ensure that the system reflects current knowledge. This external data can then be incorporated into the KG to keep the graph updated at periodic intervals. Unlike static memory systems like LLMs, which rely on outdated knowledge, NesyAI employs advanced temporal reasoning to manage static knowledge—long-established facts that remain constant—and dynamic knowledge—rapidly evolving information such as regulatory changes, scientific discoveries, and current events. This capability allows models to effectively distinguish between information requiring regular updates and data that remains relevant over time.

By leveraging temporal reasoning, NesyAI\ frameworks continuously evaluate the relevance and timeliness of the information they process. For example, these systems can identify whether legal precedents, medical guidelines, or policy updates have been superseded, adjusting their outputs accordingly. This ensures that LLMs deliver accurate, up-to-date attributions, eliminating the risk of propagating outdated or incorrect information.

Additionally, NesyAI\ frameworks prioritize recent, high-importance data over less relevant information. The dynamic memory not only adapts to real-time updates but also assesses the significance of data based on its temporal context. This ensures that citations and references are always current, reliable, and contextually appropriate, offering robust, trustworthy outputs in knowledge-intensive and fast-changing fields.

\subsection{\textbf{Metacognitive Layers for Enhanced Attribution}}
A crucial component of NesyAI is metacognition, which allows the system to intelligently decide when and how to combine neural networks with symbolic reasoning for more precise and reliable attributions. Unlike a simple step-by-step process, this interaction involves a continuous, real-time monitoring mechanism where metacognition governs the interplay between the two. While neural networks (such as LLMs) can generate rapid responses based on probabilistic models, they are prone to hallucination—regardless of the range of data they are trained on—because of the inherent uncertainty in language models. Metacognition, therefore, serves as a control layer that decides when to switch to symbolic reasoning when higher accuracy or verification is required \cite{ganapini2022thinking}.

In real time, the metacognitive layer continuously tracks key metrics such as confidence scores, ambiguity, and reliability. When the LLM generates content that falls below a certain confidence threshold or detects uncertainty in the attributions, metacognition triggers the symbolic reasoning system to cross-check the generated content against validated knowledge sources (such as domain-specific databases or KGs). While KGs are not separate systems, they serve as crucial structured repositories of verified knowledge, used to ground the model's attributions more securely.

For example, when generating legal advice or medical recommendations—domains where precision is crucial—the metacognitive layer detects when the LLM’s response may not be sufficiently reliable. It can then prompt the system to consult structured sources like a legal database or a medical KG to ensure that the information is up-to-date and accurate. Rather than applying symbolic reasoning only when the LLM output is unclear, the system could invoke symbolic reasoning every time a response is generated, ensuring that all outputs are grounded in verified data, regardless of the LLM’s confidence level. This is important because, even though the LLM may have been trained on relevant data, it can still hallucinate due to the probabilistic nature of its language generation model.

The key role of metacognition here is not to directly assess the quality of the generated content but to act as an efficient controller that decides when and how to invoke symbolic reasoning based on task complexity, content reliability, and the model's confidence. By managing this real-time interaction, the metacognitive layer ensures that hallucinations are minimized and that every attribution is grounded in a verifiable source, thus significantly enhancing the reliability of LLM-generated content, especially in high-stakes fields where inaccuracies or unverified information can have severe consequences.

\begin{figure*}[ht!] 
    \centering
    \includegraphics[width=0.9\textwidth]{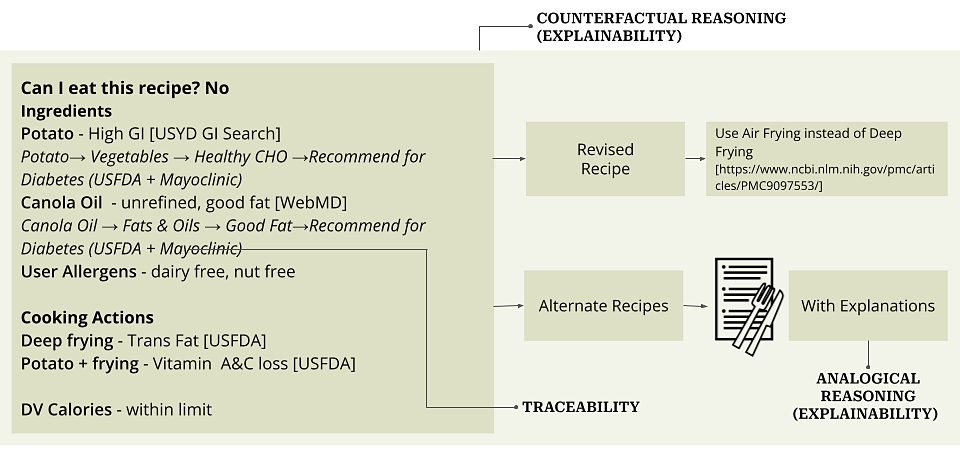} 
    \caption{An analysis of the recipe french fries in the context of diabetes with explainability and traceability. Explanations are provided in the form of reasoning. Given a recipe is deemed unsuitable, alternative suggestions are provided as a similar recipe or revising the existing recipe. Given a recipe is unsuitable, the alternative suggestions provided with explanations are called counterfactual reasoning. Explaining why an alternative recipe is similar to a given recipe refers to analogical reasoning.}
    \label{fig:food_example}
\end{figure*}

\subsection{\textbf{Explainability and Traceability Through Symbolic AI}}

A significant advantage of NesyAI is its ability to make decision-making processes both explainable and traceable back to its knowledge sources. Unlike LLMs and other neural networks, which often function as black boxes, symbolic AI offers transparent reasoning, allowing it to explain how decisions are made. This capability is crucial for attribution systems, as users must be able to trace any generated statement back to its source and understand its reasoning. For example, consider a diet management chatbot that answers questions about recipes and their relevance to dietary restrictions such as allergens or health conditions such as diabetes or hypertension. When the chatbot generates an answer using its generative model capabilities, it should also provide reasoning for its answers.

\textit{Explainability:} In the context of diabetes, both broccoli and potato are healthy carbohydrates and can be considered healthy as per Mayo Clinic (knowledge-source-1). If the glycemic index (knowledge-source-2) is factored in, potato is better avoided as it has a high glycemic index compared to broccoli. This is not only a crucial factor in recipe analysis but also in providing ingredient substitutions. If the model factored in only knowledge-source-1, it would have recommended both broccoli and potato with the same confidence. If the model factored in both knowledge-source-1 and knowledge-source-2, it would recommend broccoli over potato. Without reasoning and explanations, the user might not be able to make informed decisions, leading to adverse outcomes. In such high-stakes domains, the model needs to explain its answers. In these cases, a NesyAI model will be able to integrate multiple knowledge sources and explain its decision-making process, as shown in Figure \ref{fig:food_example}.

\textit{Traceability:} While the chatbot provides reasoning, it should also be able to support the reasoning with trusted knowledge sources. Hallucinations, giving non-sensical yet believable answers, is a well-recognized problem in LLMs. Current LLMs produce reasoning as an intrinsic part of their general answer-generation strategy without an awareness that generating reasoning necessitates accountability. NesyAI models can be well aware of accountability with respect to reasoning by providing traceable explanations with references to trusted knowledge sources. The knowledge source and the paths of decision-making can be traced for \textit{Potato} concerning diabetes as follows:

$$
\begin{array}{rl}
\text{Potato} &\rightarrow \text{Vegetables (USFDA)} \\

                   &\rightarrow \text{Healthy Carbohydrate (MayoClinic)} \\
                   &\rightarrow \text{Recommend (MayoClinic)} 
\end{array}
$$

$$
\begin{array}{rl}
\text{Potato} &\rightarrow \text{Glycemic Index:72 (GI Search DB)} \\

                   &\rightarrow \text{Recommend with Caution (MayoClinic)} \\ 
\end{array}
$$

As per USFDA, Potato belong to the vegetables category which is considered a healthy carbohydrate by MayoClinic and so it can be recommended to the user. When we factor in another context, glycemic index, MayoClinic suggests to moderate intake of foods with high glycemic index. This shows the importance of explanations and reasoning which aids the user in informed decision making. Figure \ref{fig:food_example} also shows different kinds of explanations provided in the form of reasoning along with traceability. Given a recipe is unsuitable, the alternative suggestions provided with explanations are called counterfactual reasoning. Explaining why an alternative recipe is similar to a given recipe refers to analogical reasoning \footnote{https://www.youtube.com/watch?v=XDQTxorwypo}.

\subsection{\textbf{Dynamic Knowledge Update for Continual Learning}}
One of the significant shortcomings of current attribution systems is their inability to manage rapidly changing information. In contrast, NesyAI employs a more structured approach to handle dynamic updates. While KGs serve as the foundational component for symbolic reasoning within NesyAI, they are traditionally curated by domain experts, ensuring their information is validated and reliable. However, this expert curation introduces an inherent limitation regarding real-time adaptability, as updating KGs typically involves delays due to validation cycles. To overcome this, NesyAI incorporates a hybrid solution where the KG remains the trusted source for verified, long-standing knowledge. In contrast, dynamic external data sources, such as real-time databases or streaming APIs, are integrated to provide updates on rapidly evolving information. These dynamic sources in an RAG system enable the system to temporarily adapt to new developments, allowing it to respond in near real-time to changes in domains like news, medicine, and law, where information constantly evolves. This layered approach enables NesyAI to provide immediate, real-time responsiveness while awaiting formal updates to the KG by domain experts. By blending the reliability of the KG with the adaptability of real-time data, NesyAI ensures that its attributions remain relevant and accurate, even in fast-changing information landscapes. This approach preserves the immediacy required for up-to-date information and the precision of expert-validated knowledge.
In \autoref{tab:comparison}, we summarize the main differences between current approaches for attribution and NesyAI, highlighting how this hybrid system overcomes the limitations of traditional methods.

\begin{table*}[ht]
\centering
\caption{Comparative Analysis of Attribution Methods in Traditional vs. Neurosymbolic AI: Addressing Limitations in Accuracy, Transparency, and Real-Time Adaptability.}
\begin{tabular}{|p{2.8cm}|p{5cm}|p{5cm}|}
\hline
\textbf{Aspect}                        & \textbf{Current LLM-Based Attribution}                                                        & \textbf{Proposed Neurosymbolic Attribution}                                         \\ \hline
\textbf{Hallucination Reduction}        & LLMs frequently generate factually incorrect or unsupported content (hallucinations), with limited ability to cross-check for verifiable sources. & Reduces hallucinations by grounding outputs in structured, verifiable sources using KGs and symbolic reasoning. \\ \hline
\textbf{Transparency and Explainability} & Current systems function as black boxes, with limited transparency about how sources are selected or why certain content is generated. & Offers full traceability and clear reasoning paths, allowing users to understand and verify each decision made during content generation. \\ \hline
\textbf{Dynamic Knowledge Updating}     & Relies on static datasets, making it difficult to update citations in real-time or adjust to evolving knowledge.                      & Supports dynamic, real-time updates through the integration of KGs, ensuring that attributions remain current and relevant. \\ \hline
\textbf{Legal and Ethical Compliance}   & High risk of copyright infringement and improper attribution of sources, often lacking compliance with intellectual property laws.                      & Embeds legal ontologies and ensures proper attribution of sources, adhering to copyright and intellectual property regulations. \\ \hline
\textbf{Precision and Source Relevance} & LLMs often provide irrelevant, outdated, or low-quality citations, which can reduce the trustworthiness and precision of the output.                           & Enhances precision by selecting relevant, high-quality, peer-reviewed sources from structured knowledge repositories.    \\ \hline
\end{tabular}
\label{tab:comparison}
\end{table*}

\section{Challenges and Limitations}

One of the primary challenges in current attribution models is the lack of transparency. Current LLMs often do not provide clear explanations of how they arrived at a specific output or why a particular source was cited. This black-box nature has led to a growing distrust of AI-generated outputs, especially in legal and academic contexts where the traceability of sources is critical. Without a clear understanding of the reasoning behind citations, users are left unsure of the credibility and accuracy of the information provided.

The main challenges in current attribution models can be itemized as follows:

\begin{itemize}
    \item \textbf{Lack of Transparency}: LLMs often function as black boxes, where the reasoning behind selecting a source or the generation process of outputs remains unclear. This lack of explainability erodes trust in AI-generated content, especially in high-stakes domains like law and academia.
    
    \item \textbf{Outdated Knowledge}: LLMs struggle to keep pace with rapidly evolving information landscapes. The sources used in generating responses may quickly become outdated, making it difficult for models to provide up-to-date and relevant attributions, especially in dynamic fields like healthcare and law.
    
    \item \textbf{Proliferation of AI-generated Data}: The exponential increase in AI-generated data poses a significant challenge. AI systems often use data generated by other AI models, creating a loop where AI-produced content is recycled without proper attribution or verification of original sources. This leads to greater difficulty in tracing the provenance of information and increases the risk of misinformation and errors.
    
    \item \textbf{Legal and Ethical Compliance}: Integrating legal requirements into LLMs is challenging. The Digital Millennium Copyright Act (DMCA) offers limited protection for authors whose works are used in training datasets. Furthermore, the absence of a straightforward mechanism to enforce the right to attribution in generative AI increases the complexity of ensuring compliance with intellectual property laws. The proliferation of AI-generated data adds another layer of difficulty in distinguishing between human-generated and AI-generated content.
\end{itemize}

Given these limitations, there is a clear need for advanced attribution systems that comply with both ethical and legal standards. The rapidly growing use of AI-generated data, combined with the lack of transparency, outdated knowledge, and legal challenges, makes it imperative to develop systems that can provide accurate, up-to-date, and verifiable attributions.

\section{Current Progress and Future Directions}

The issue of attribution in LLMs has gained significant attention due to technical limitations and legal and ethical concerns regarding the use of copyrighted material without proper attribution. Despite advancements, current attribution methods fail to ensure reliability, transparency, and compliance with legal frameworks, especially in high-stakes environments like healthcare and law. Different types of attribution have been explored, each contributing incremental improvements but also exhibiting limitations:

\begin{itemize}
    \item \textbf{Post-Generation Attribution:} One of the earliest methods developed was post-generation attribution, where the model generates text first and then attempts to find sources that match the generated content. This technique has been central to RAG, which retrieves external data to generate answers. While this approach improves the ability to provide some form of citation, it often struggles with precision, as the model may match incorrect or irrelevant sources post hoc. The lack of a structured approach during content generation means that hallucinations—factually incorrect or unsupported content—are still prevalent. Systems like OpenAI’s GPT models often generate content first, and then attempt to cite sources, but this can lead to fabricated or mismatched references. In many cases, users cannot trace the exact reasoning behind why a particular source was chosen.
    \item \textbf{Inline (During-Generation) Attribution:} As an improvement over post-generation methods, inline attribution provides citations as the model generates content. The goal here is to enable the LLM to link specific pieces of information directly to sources in real-time, which can improve precision and reduce hallucinations. However, this approach is computationally expensive and requires extensive integration with real-time knowledge sources. Additionally, current models struggle to efficiently balance the needs of citation accuracy with the fluidity of natural language generation. Inline attribution, although more accurate, faces limitations when LLMs attempt to handle open-ended queries that require dynamic knowledge retrieval.

    \item \textbf{Pre-Generation Attribution:} To address the limitations of post and inline attribution models, some LLMs utilize pre-generation attribution, where sources are retrieved before generating content. This method ensures that content is generated based on pre-verified, structured information, significantly reducing the risk of hallucinations. Pre-generation techniques have been explored in knowledge-intensive applications where factual accuracy is critical, such as scientific paper generation and legal document synthesis. Despite this improvement, these methods may limit the flexibility of the model by constraining it to predefined data. Google’s Bard AI, which attempts to retrieve relevant information before generating text, aims to improve precision in high-stakes environments like healthcare and law but can suffer from limited adaptability to unforeseen or evolving queries.

    \item \textbf{Hybrid Attribution:} Some of the most recent developments in attribution combine aspects of both inline and post-generation approaches, known as hybrid attribution. Hybrid models dynamically adjust between real-time source retrieval during content generation and post-generation verification, aiming to provide both flexibility and precision. This method holds promise for balancing accuracy with the computational efficiency needed for broader, real-world deployment, but it remains in the experimental stages. These hybrid models are currently being explored for tasks requiring dynamic updates in knowledge-heavy fields such as academic research, where both accuracy and adaptability are critical.
\end{itemize}
A NesyAI framework to enhance attribution in AI systems, using LLMs as a prime example, addressing the limitations of traditional methods. Integrating neural adaptability with symbolic reasoning significantly reduces hallucinations and improves the transparency of outputs. KGs further enrich this integration, providing essential real-world context and ensuring that the system complies with ethical standards and user expectations. NesyAI framework offers a robust solution to the challenges of attribution, aligning AI-generated content with both factual accuracy and legal requirements.

\section{Acknowledgments}
This research was supported by NSF Award \#2335967, ``EAGER: Knowledge-guided Neurosymbolic AI with Guardrails for Safe Virtual Health Assistants." The opinions expressed are those of the authors and do not necessarily reflect the views of the NSF. We acknowledge Vedant Khandelwal for providing valuable feedback on earlier drafts.
\section{Authors}

\textbf{Deepa Tilwani} is a Ph.D. student at the AIISC, with a research focus on developing neurosymbolic AI, and its applications to healthcare. Contact her at: \url{dtilwani@mailbox.sc.edu.}

\noindent\textbf{Revathy Venkataramanan} is a Ph.D. student at the AIISC, with a research focus on explainable AI models with reasoning and traceability, multi-contextual grounding and process knowledge graphs. Contact her at: \url{revathy@email.sc.edu.}

\noindent \textbf{Amit P. Sheth} is the NCR Chair, and a professor; he founded the university-wide AI Institute of South Carolina (AIISC) in 2019. He received the 2023 IEEE-CS Wallace McDowell award and is a fellow of IEEE, AAAI, AAIA, AAAS, and ACM. Contact him at: \url{amit@sc.edu}

\bibliographystyle{IEEEtran}
\bibliography{ref}

\begin{thebibliography}{10}
\providecommand{\url}[1]{#1}
\csname url@samestyle\endcsname
\providecommand{\newblock}{\relax}
\providecommand{\bibinfo}[2]{#2}
\providecommand{\BIBentrySTDinterwordspacing}{\spaceskip=0pt\relax}
\providecommand{\BIBentryALTinterwordstretchfactor}{4}
\providecommand{\BIBentryALTinterwordspacing}{\spaceskip=\fontdimen2\font plus
\BIBentryALTinterwordstretchfactor\fontdimen3\font minus \fontdimen4\font\relax}
\providecommand{\BIBforeignlanguage}[2]{{%
\expandafter\ifx\csname l@#1\endcsname\relax
\typeout{** WARNING: IEEEtran.bst: No hyphenation pattern has been}%
\typeout{** loaded for the language `#1'. Using the pattern for}%
\typeout{** the default language instead.}%
\else
\language=\csname l@#1\endcsname
\fi
#2}}
\providecommand{\BIBdecl}{\relax}
\BIBdecl

\bibitem{Xu2024HallucinationII}
Z.~Xu, S.~Jain, and M.~Kankanhalli, ``Hallucination is inevitable: An innate limitation of large language models,'' \emph{arXiv preprint arXiv:2401.11817}, 2024.

\bibitem{peskoff-stewart-2023-credible}
\BIBentryALTinterwordspacing
D.~Peskoff and B.~Stewart, ``Credible without credit: Domain experts assess generative language models,'' in \emph{Proceedings of the 61st Annual Meeting of the Association for Computational Linguistics (Volume 2: Short Papers)}, A.~Rogers, J.~Boyd-Graber, and N.~Okazaki, Eds.\hskip 1em plus 0.5em minus 0.4em\relax Toronto, Canada: Association for Computational Linguistics, Jul. 2023, pp. 427--438. [Online]. Available: \url{https://aclanthology.org/2023.acl-short.37}
\BIBentrySTDinterwordspacing

\bibitem{Li2023ASO}
D.~Li, Z.~Sun, X.~Hu, Z.~Liu, Z.~Chen, B.~Hu, A.~Wu, and M.~Zhang, ``A survey of large language models attribution,'' \emph{arXiv preprint arXiv:2311.03731}, 2023.

\bibitem{Malaviya2023ExpertQAEQ}
C.~Malaviya, S.~Lee, S.~Chen, E.~Sieber, M.~Yatskar, and D.~Roth, ``Expertqa: Expert-curated questions and attributed answers,'' \emph{arXiv preprint arXiv:2309.07852}, 2023.

\bibitem{jayakumar-etal-2023-large}
\BIBentryALTinterwordspacing
T.~Jayakumar, F.~Farooqui, and L.~Farooqui, ``Large language models are legal but they are not: Making the case for a powerful {L}egal{LLM},'' in \emph{Proceedings of the Natural Legal Language Processing Workshop 2023}, D.~Preo{\textcommabelow{t}}iuc-Pietro, C.~Goanta, I.~Chalkidis, L.~Barrett, G.~Spanakis, and N.~Aletras, Eds.\hskip 1em plus 0.5em minus 0.4em\relax Singapore: Association for Computational Linguistics, Dec. 2023, pp. 223--229. [Online]. Available: \url{https://aclanthology.org/2023.nllp-1.22}
\BIBentrySTDinterwordspacing

\bibitem{Liu2023EvaluatingVI}
N.~F. Liu, T.~Zhang, and P.~Liang, ``Evaluating verifiability in generative search engines,'' \emph{arXiv preprint arXiv:2304.09848}, 2023.

\bibitem{Tilwani2024REASONSAB}
D.~Tilwani, Y.~Saxena, A.~Mohammadi, E.~Raff, A.~Sheth, S.~Parthasarathy, and M.~Gaur, ``Reasons: A benchmark for retrieval and automated citations of scientific sentences using public and proprietary llms,'' \emph{arXiv preprint arXiv:2405.02228}, 2024.

\bibitem{li2023towards}
X.~Li, Y.~Cao, L.~Pan, Y.~Ma, and A.~Sun, ``Towards verifiable generation: A benchmark for knowledge-aware language model attribution,'' \emph{arXiv preprint arXiv:2310.05634}, 2023.

\bibitem{bohnet2022attributed}
B.~Bohnet, V.~Q. Tran, P.~Verga, R.~Aharoni, D.~Andor, L.~B. Soares, M.~Ciaramita, J.~Eisenstein, K.~Ganchev, J.~Herzig \emph{et~al.}, ``Attributed question answering: Evaluation and modeling for attributed large language models,'' \emph{arXiv preprint arXiv:2212.08037}, 2022.

\bibitem{Asai2024ReliableAA}
A.~Asai, Z.~Zhong, D.~Chen, P.~W. Koh, L.~Zettlemoyer, H.~Hajishirzi, and W.-t. Yih, ``Reliable, adaptable, and attributable language models with retrieval,'' \emph{arXiv preprint arXiv:2403.03187}, 2024.

\bibitem{Sheth2023NeurosymbolicA}
A.~Sheth, K.~Roy, and M.~Gaur, ``Neurosymbolic artificial intelligence (why, what, and how),'' \emph{IEEE Intelligent Systems}, vol.~38, no.~3, pp. 56--62, 2023.

\bibitem{Gaur2023BuildingTN}
M.~Gaur and A.~Sheth, ``Building trustworthy neurosymbolic ai systems: Consistency, reliability, explainability, and safety,'' \emph{AI Magazine}, vol.~45, no.~1, pp. 139--155, 2024.

\bibitem{goonmeetGrounding}
G.~K. Bajaj, V.~Shalin, S.~Parthasarathy, and A.~Sheth, ``Grounding from an ai and cognitive science lens,'' \emph{IEEE Intelligent Systems}, vol.~39, pp. 66--71, 03 2024.

\bibitem{Adam2024TraceableLV}
D.~Adam and T.~Kliegr, ``Traceable llm-based validation of statements in knowledge graphs,'' \emph{arXiv preprint arXiv:2409.07507}, 2024.

\bibitem{ganapini2022thinking}
M.~B. Ganapini, M.~Campbell, F.~Fabiano, L.~Horesh, J.~Lenchner, A.~Loreggia, N.~Mattei, F.~Rossi, B.~Srivastava, and K.~B. Venable, ``Thinking fast and slow in ai: The role of metacognition,'' in \emph{International Conference on Machine Learning, Optimization, and Data Science}.\hskip 1em plus 0.5em minus 0.4em\relax Springer, 2022, pp. 502--509.

\end{thebibliography}

\end{document}